\RequirePackage{amsmath} 
\documentclass[runningheads]{llncs}
\usepackage{amsfonts}
\usepackage{amssymb}
\usepackage{mathrsfs}
\usepackage{subfigure}
\usepackage{pstricks}
\usepackage{float}
\usepackage{appendix}
\usepackage{overpic}
\usepackage{enumerate}
\usepackage[colorlinks,linkcolor=black,citecolor=blue,urlcolor=blue]{hyperref}
\usepackage[UKenglish]{babel}
\usepackage{comment}
\usepackage{pifont}
\usepackage{graphicx}
\usepackage{color}
\usepackage{hyperref}
\usepackage{tikz}
\usetikzlibrary{positioning} 
\usepackage{xcolor}
\usepackage{graphicx}
\usepackage{comment}
\usepackage{marvosym}
\newtheorem{defn}{Definition}
\newtheorem{thm}{Theorem}
\newtheorem{lem}{Lemma}
\newtheorem{exmp}{Example}

\newtheorem{prop}{Proposition}
\newtheorem{post}{Postulate}

\begin{document}
\title{Fuzzy Labeling Semantics for Quantitative Argumentation}
%
\author{Zongshun Wang\and
Yuping Shen\textsuperscript{{(}\Letter{)}}}
\authorrunning{Z. Wang and Y. Shen}
%
\institute{Institute of Logic and Cognition, \\
Department of Philosophy, \\Sun Yat-sen University, Guangzhou, China\\
\email{wangzsh7@mail2.sysu.edu.cn}, \email{shyping@mail.sysu.edu.cn}
}
\maketitle           

\begin{abstract}
Evaluating \emph{argument strength} in  quantitative argumentation systems has received increasing attention in the field of abstract argumentation.
The concept of  \emph{acceptability degree} is widely adopted in \emph{gradual semantics}, however, it may not be sufficient in many practical applications. In this paper, we provide a novel quantitative method called \emph{fuzzy labeling} for \emph{fuzzy argumentation systems}, in which  a \emph{triple} of \emph{acceptability}, \emph{rejectability}, and \emph{undecidability} \emph{degrees} is used to evaluate argument strength. 
Such a setting sheds new light on defining argument strength and provides a deeper understanding of the status of arguments. More specifically, we investigate the \emph{postulates} of fuzzy labeling, which present the rationality requirements for semantics concerning the acceptability, rejectability, and undecidability degrees.
We then propose a class of fuzzy labeling semantics conforming to the above postulates and investigate the relations between fuzzy labeling semantics and existing work in the literature.

\end{abstract}

\section{Introduction}

The theory of \emph{abstract argumentation} was first proposed in Dung's seminal paper \cite{dung1995acceptability} and now plays an important role in artificial intelligence \cite{bench2007argumentation}.
The fundamental idea of abstract argumentation theory is \emph{argumentation framework} (AF), which is essentially a directed graph whose nodes represent \emph{arguments} and arrows represent \emph{attack relation} between arguments.

In recent years, the study of \emph{quantitative argumentation systems} (QuAS) has received increasing attention and numerous QuAS have been defined via different quantitative approaches, such as \emph{weighted argumentation systems} (WAS) \cite{Bistarelli2021weighted,dunne2011weighted},
\emph{probabilistic argumentation systems} (PAS) \cite{hunter2013probabilistic,hunter2021probabilistic,li2011probabilistic}, 
\emph{fuzzy argumentation systems} (FAS) \cite{da2011changing,janssen2008fuzzy,tamani2014fuzzy,wu2016godel}, etc.
Generally speaking, each argument or attack in a QuAS is assigned an \emph{initial degree}, usually expressed by a \emph{numerical value} in $[0,1]$ from a meta-level, so that richer real-world applications can be properly described. 

In abstract argumentation theory, the evaluation of arguments is a central topic, and it is commonly achieved through \emph{semantics} \cite{baroni2011introduction}. For example, the well-known \emph{extension semantics} and \emph{labeling semantics} are designed to deal with classical AFs, giving sets of acceptable arguments and labels \{accepted, rejected, undecided\} over arguments respectively, while in QuAS the \emph{gradual semantics} are used for evaluating the \emph{strength} of arguments by assigning each argument a numerical value in $[0,1]$ as the so-called \emph{acceptability degree}.

The study of gradual semantics has received extensive attention in the literature \cite{Amgoud2021Evaluation,baroni2019fine,Besnard2001Logic,beuselinck2023principle,cayrol2005graduality,oren2022inverse}.
Most of the work focuses on the evaluation of the acceptability degree. However, we argue that this approach may not always be sufficient in practical applications. 
To make more informed decisions, a rational agent may need to evaluate both positive and negative aspects, as evidenced by a body of literature in many research areas \cite{Amgoud2009using,bouzarour2015bipolar,cacioppo1994relationship,Dubois2006Qualitative,Leite2011Social,osgood1957measurement,Walton2009Explanations,haenni2009probabilistic}. 
While the acceptability degree measures the extent to which an argument can be accepted (reflecting its positive aspect), the impact of its attackers (reflecting its negative aspect) should also be characterized.

Motivated by the observation that an argument suffering more attack is more likely to be rejected, we propose the concept of \emph{rejectability degree}, measuring the extent to which the argument can be rejected according to the impact of its attackers. 
The rejectability degree helps to make more informed decisions, especially in cases where minimizing attack is crucial. 
For instance, politicians may prefer to choose ``safer'' arguments (i.e., suffer less attack) to avoid criticism or risks. 
In addition, we introduce the notion of \emph{undecidability degree}, which measures the extent to which the argument cannot be decided to be accepted or rejected. 
This notion allows for capturing the degree of ``uncertainty" or ``don't know'', which is widely adopted in various fields, such as Dempster-Shafer theory \cite{dempster2008dempster,shafer1976mathematical}, subjective logic \cite{josang1997artificial}, and safety-critical domain \cite{wang2019safety}.
We illustrate the above idea  with the example below.

\begin{exmp}\label{example1}
Consider the following scenario:
\begin{itemize}
    \item A: Getting vaccinated may cause side effects.
    \item  B: Everyone should get vaccinated due to the viral pandemic.
\end{itemize}

This instance is represented as a QuAS in Figure \ref{figure 1}, in which $A$ attacks $B$, $A$ and $B$ are assigned initial degrees $0.3$ and $1$ respectively.

\begin{figure}[h]\label{figure 1}
\centering
\begin{tikzpicture}[
roundnode/.style={circle, draw=black!100, fill=white!5, thick, minimum size=5mm},
rectanglenode/.style={rectangle, draw=black!100, fill=white!5, thick, minimum size=0mm},
]
\node[roundnode]      (1) at(-2,0)      {$A_{0.3}$};
\node[roundnode]      (2) at(2,0)      {$B_{1.0}$};
\draw[->,thick] (1)--(2);
\end{tikzpicture}
\caption{Getting Vaccinated or Not}
\end{figure}
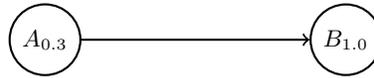

We analyze the strength of $A$ and $B$ as follows (shown in Figure \ref{figure 2.1}).
\begin{enumerate}
\item The acceptability degree of $A$ remains its initial degree $0.3$ since $A$ has no attackers, while the acceptability degree of  $B$ is $1-0.3=0.7$, i.e., obtained from weakening its initial degree through attacker $A$.
\item The rejectability degree of $A$ is $0$ since it  has no attackers. The rejectability degree of $B$ is $0.3$ since its  attacker  $A$ has acceptability degree $0.3$, and  it is reasonable to reject $B$ to the same extent.
\item The undecidability degree of $A$ is $1-0.3=0.7$ and $B$ is $1-(0.7+0.3)=0$.
\end{enumerate}
\begin{figure}[h]
\centering
\begin{tabular}{l|cccc}
 & $A$ & & & $B$\\
 \hline
acceptability degree & $0.3$ & & & $0.7$\\
  \hline
rejectability degree & $0$ & & & $0.3$\\
 \hline
undecidability degree & $0.7$& & & $0$\\
\hline
\end{tabular}
\caption{The Extended Argument Strength of $A$ and $B$}
\label{figure 2.1}\end{figure}

Existing evaluation methods suggest that argument $B$ is preferable to $A$ due to its higher acceptability degree $(0.7>0.3)$. However, in the real world, many people choose not to get vaccinated due to potential side effects, i.e., prefer $A$ to $B$.
Our approach suggests that argument $A$ is preferable to $B$ due to its lower rejectability degree $(0<0.3)$.
So for these people with safety concerns, the rejectability degree appears more critical to avoid risks.
\end{exmp}

In the paper, we propose a more comprehensive evaluation method called \emph{fuzzy labeling} which describes argument strength as a triple consisting of acceptability, rejectability and undecidability degrees. 
In essence, fuzzy labeling is a combination of the gradual semantics and labeling semantics, by assigning a numerical value to each label \{accepted, rejected, undecided\}.
Such a setting provides new insights into argument strength and a deeper understanding of the status of arguments.
Due to its expressiveness and flexibility, fuzzy labeling is suitable for many potential applications, e.g. in engineering control where reliability is a major concern and thus minimizing attacks is necessary to avoid risks or costs. 
Furthermore, it is beneficial to identify the status of rejected and undecided in judgment aggregation \cite{caminada2011judgment,caminada2011manipulation}, designing algorithms \cite{caminada2007algorithm,cerutti2014scc}, explaining semantics \cite{schulz2018responsibility}, etc.

After introducing the framework, we propose a class of
\emph{fuzzy labeling semantics} by using the well-known \emph{postulate-based approach} \cite{van2017principle} in two steps: 
(i) Investigate the \emph{postulates} for fuzzy labeling, which adapt the criteria for classical labeling semantics \cite{caminada2009logical} and incorporate the concept of \emph{tolerable attack} \cite{wu2016godel};
(ii) Formalize fuzzy labeling semantics that conform to the above postulates.
Finally, we discuss the relationships between fuzzy labeling semantics and existing work, including classical labeling semantics \cite{caminada2009logical}, fuzzy extension semantics \cite{wu2016godel}, etc. 

The remaining part of this paper is structured as follows. We recall basic concepts in Section \ref{preliminaries} and define fuzzy labeling semantics  in Section \ref{section fuzzy labeling Semantics}.
In Section \ref{section relationship} we discuss the relations between fuzzy labeling semantics and related work. The paper ends with conclusions and remarks about future work.


\section{Preliminaries}\label{preliminaries}

\subsection{Fuzzy Set Theory}
\begin{defn}[\cite{zadeh1965fuzzy}]
A \emph{fuzzy set} is a pair $(X, S)$ in which $X$ is a nonempty set called the \emph{universe} and $S: X\rightarrow[0, 1]$ is the associated \emph{membership function}. For each $x\in X$,  $S(x)$ is called the \emph{grade} of membership of $x$ in $X$. \end{defn} 

For convenience, when the universe $X$ is fixed,  a fuzzy set  $(X, S)$ is identified by its membership function $S$,  which can be represented by  a set of pairs $(x,a)$ with $x\in X$ and $a\in[0,1]$. We stipulate that all pairs $(x,0)$ are omitted from $S$.

For instance, the following are fuzzy sets with universe $\{A,B,C\}$: 
$$S_1=\{(A,0.5)\},\  S_2=\{(B,0.8),(C,0.9)\},\  S_3=\{(A,0.8),(B,0.8),(C,1)\}.$$
Note that $S_1(A)=0.5, S_1(B)=S_1(C)=S_2(A)=0$, and in $S_3$ every element has a non-zero grade.

A \emph{fuzzy point} is a fuzzy set containing a unique pair $(x,a)$. We may identify a fuzzy point  by its pair. For example, $S_1$ is a fuzzy point and identified by $(A,0.5)$.

Let $S_1$ and $S_2$ be two fuzzy sets. Say $S_1$ is a \emph{subset} of $S_2$, denoted by $S_1\subseteq S_2$, if for any $x\in X$, $S_1(x)\le S_2(x)$. Conventionally, we write $(x, a)\in S$ if a fuzzy point  $(x, a)$ is a subset of $S$. Moreover, we shall use the following notations:
\begin{itemize}
\setlength{\itemsep}{0pt}
\setlength{\parsep}{0pt}
\setlength{\parskip}{0pt}
  \item the \emph{union}  of $S_1$ and $S_2$: $S_1\cup S_2=\{(x,\max\{S_1(x),S_2(x)\})\mid x\in X\}$;
  \item the \emph{intersection} of $S_1$ and $S_2$: $S_1\cap S_2=\{(x,\min\{S_1(x),S_2(x)\})\mid x\in X\}$;
  \item the \emph{complement}  of $S$: $S^c=\{(x,1-a)\mid S(x)=a\}$;
\end{itemize}

In this example,  $S_1(x)\leq S_3(x)$ for each element $x$, thus fuzzy point $S_1$ is a subset of $S_3$, written as $(A,0.5)\in S_3$. Similarly, it is easy to check: (i) $S_2\subseteq S_3$; (ii)  $S_2\cup S_3=\{(A,0.8),(B,0.8),(C,1)\}$; (iii) $S_1\cap S_3=\{(A,0.5)\}$; (vi) $S_3^c=\{(A,0.2),(B,0.2)\}$. 

\subsection{Fuzzy Argumentation System}\label{the preliminaries of FAS}

\emph{Fuzzy argumentation system} (FAS) \cite{da2011changing,janssen2008fuzzy,wu2016godel} extends classical argumentation framework with fuzzy degree on arguments and attacks.
Each argument or attack has an initial degree from the interval $[0,1]$.
The initial degree is usually assigned from a meta-level, and we simply assume that initial degrees are pre-assigned.


\begin{defn}
\label{FAS}
A fuzzy argumentation system over a finite set of arguments $Args$ is a pair  $\mathcal{F}=\langle\mathcal{A}, \mathcal{R}\rangle$ in which $\mathcal{A}: Args \rightarrow [0, 1]$ and $\mathcal{R}: Args \times Args \rightarrow [0, 1]$ are total functions. 
\end{defn}

In Definition \ref{FAS}, $\mathcal{A}$ and  $\mathcal{R}$ are fuzzy sets of arguments and attacks. $\mathcal{A}$ can be denoted by pairs $(A,\mathcal{A}(A))$ in which $\mathcal{A}(A)$ is the initial degree of $A$, and $\mathcal{R}$ can be denoted by pairs $\big( (A,B),\mathcal{R}(A,B)\big)$ or simply $\big( (A,B),\mathcal{R}_{_{AB}}\big)$. Moreover, we denote by $Att(A)$ the set of all  attackers of $A$, i.e.,  $Att(A)=\{B\in Args\mid  \mathcal{R}_{_{BA}}\neq 0\}.$ 

In Definition \ref{attack intensity}, we define the \emph{attack intensity} to show the impact of attackers on the attacked arguments.

\begin{defn}\label{attack intensity}
Let  $\langle\mathcal{A}, \mathcal{R}\rangle$ be an FAS and $A,B\in Args$. We define that
\begin{itemize}
    \item the \emph{attack intensity} of $(B,b)\in \mathcal{A}$ towards $A$ w.r.t. $\mathcal{R}_{_{BA}}$ is $b*\mathcal{R}_{_{BA}}$,
    \item  the \emph{attack intensity} of  $\mathcal{S}\subseteq\mathcal{A}$ towards  $A$ is $\max_{B\in Att(A)} \mathcal{S}(B)*\mathcal{R}_{_{BA}}$,
\end{itemize}
where $*$ is a binary operator s.t. $a*b=\min\{a,b\}$.\footnote{For simplicity, we adopt the operation `min' in this paper, and it can be extended to other operations, such as product and Lukasiewicz, for real-world applications.}
\end{defn}

Unlike Dung's semantics, in the semantics of FAS (or other QuAS), two (fuzziness) arguments with a weak attack relation can be accepted together. 
Namely, it allows for a certain degree of tolerance towards attacks between arguments.
In this paper, we adopt the notion of \emph{tolerable attack} introduced in \cite{wu2016godel}: an attack is considered \emph{tolerable} if the sum of the attacker's attack intensity and the attackee's degree is not greater than 1. 
An appealing property is that their semantics, defined within this setting, are compatible with Dung's admissibility semantics (Section 5 of \cite{wu2016godel}). 
This compatibility is useful in our application of fuzzy labeling to generalize classical semantics.

%

\begin{defn} \label{sufficient attack}
Let $\langle\mathcal{A}, \mathcal{R}\rangle$ be an FAS,  $(A, a), (B, b)\in \mathcal{A}$ and $\big( (A, B), \mathcal{R}_{_{AB}}\big)\in \mathcal{R}$. If $a*\mathcal{R}_{_{AB}}+b\leq 1$, then the attack from $(A, a)$ to $(B, b)$ is called  \emph{tolerable}, otherwise it is called \emph{sufficient}.
\end{defn}

\begin{exmp}[Cont.]\label{example2}
Consider the FAS $\langle\mathcal{A},\mathcal{R}\rangle$ depicted in Example \ref{example1},
in which  $\mathcal{A}=\{(A,0.3),(B,1)\}$, and  $\mathcal{R}=\{\big( (A,B),1\big)\}$.
We directly obtain that the initial degree of $A$ is $0.3$ and $B$ is $1$, $Att(A)=\varnothing$ and $Att(B)=\{A\}$.
Moreover, the attack intensity of $(A,0.3)$ towards $B$ w.r.t. $\big( (A,B),1\big)$ is $0.3*1=\min\{0.3,1\}=0.3.$
Since $0.3*1+1>1$, the attack from $(A,0.3)$ to $(B,1)$ is sufficient.
\end{exmp}

\section{Fuzzy Labeling Semantics for FAS}\label{section fuzzy labeling Semantics}

\subsection{Fuzzy Labeling and Its Postulates}

In this section, we extend classical labeling theory in \cite{caminada2006issue} to \emph{fuzzy labeling} for FAS.
While classical labeling assigns each argument a label from \{accepted, rejected, undecided\},
fuzzy labeling assigns each argument a triple consisting of acceptability, rejectability and undecidability degrees.

\begin{defn}
Let $\mathcal{F}=\langle\mathcal{A}, \mathcal{R}\rangle$ be an FAS over a finite set of arguments $Args$. A \emph{fuzzy labeling}  for $\mathcal{F}$ is a total function $${F\!Lab}_\mathcal{F}:Args\rightarrow [0,1]\times[0,1]\times [0,1].$$ We denote $F\! Lab_\mathcal{F}(A)$ by a triple $(A^{a},A^{r},A^{u})$ where each element is  respectively called  the \emph{acceptability}, \emph{rejectability}, \emph{undecidability} \emph{degree} of argument $A$. For convenience, $F\! Lab$ can also be written as a triple $(F\!Lab^{a}_\mathcal{F}, F\!Lab^{r}_\mathcal{F},F\!Lab^{u}_\mathcal{F})$,
where each $F\! Lab^{\circ}_\mathcal{F}$ is a fuzzy set defined as  $\{(A, A^{\circ})\mid A\in Args\}$ with $\circ\in\{a,r,u\}$.
\end{defn}

When the context is clear, we will use the shorthand ${F\!Lab}$ instead of ${F\!Lab}_\mathcal{F}$.
For simplicity, we shall use \emph{acceptability} (resp.  \emph{rejectability, undecidability}) \emph{arguments} to refer to the elements in $F\!Lab^a$ (resp. $F\!Lab^r$, $F\!Lab^u$).

\begin{exmp}[Cont.]\label{example3}
Continue Example \ref{example2}. Let 
$$F\! Lab=(\{(A,0.3),(B,0.7)\},\{(B,0.3)\},\{(A,0.7)\})$$ be a fuzzy labeling for FAS.
Then $F\!Lab(A)=(0.3,0,0.7)$ and $F\!Lab(B)=(0.7,0.3,0)$. 
More precisely, the acceptability, rejectability and undecidability degree of $A$ is $0.3$, $0$ and $0.7$ respectively. Similarly, the corresponding degree of $B$ is $0.7$, $0.3$ and $0$ respectively.
\end{exmp}

We aim to use fuzzy labeling to generalize several widely studied classical semantics, including conflict-free, admissible, complete, preferred, grounded, semi-stable, and stable semantics (see \cite{baroni2011introduction} for an overview).
To achieve this, we provide a set of \emph{postulates}, each representing a rational constraint on acceptability, rejectability, or undecidability degree. 
We recall the meaning of the three degrees:
(i) the acceptability degree of an argument measures the extent to which it can be accepted,
(ii) the rejectability degree measures the extent to which it can be rejected, and 
(iii) the undecidability degree measures the extent to which it cannot be decided to be accepted or rejected. 

In the literature, the initial degree usually represents the maximal degree to which an argument can be accepted \cite{Amgoud2021Evaluation,baroni2019fine,da2011changing,wu2016godel}. So we turn to the first basic postulate, called \emph{Bounded}, which states that the acceptability degree of an argument is bounded by its initial degree.

\begin{post}[Bounded, BP]\label{BP}
A fuzzy labeling satisfies the \emph{Bounded Postulate} over an FAS $\langle\mathcal{A},\mathcal{R}\rangle$ iff  $\forall A\in Args$, 
$A^{a}\leq \mathcal{A}(A)$.
\end{post}

As shown before, the undecidability degree measures the extent to which an argument cannot be decided to be accepted or rejected.
It represents the degree of ``uncertainty'' regarding the argument.
This leads to the second basic postulate, called \emph{Uncertainty}.

\begin{post}[Uncertainty, UP]\label{UP}
A fuzzy labeling satisfies the \emph{Uncertainty Postulate}  over an FAS $\langle\mathcal{A},\mathcal{R}\rangle$ iff  $\forall A\in Args$, $A^{u}=1-A^{a}-A^{r}.$
\end{post}

In the following, we establish postulates to refine three basic semantics: conflict-free, admissible, and complete.
According to \cite{baroni2011introduction}, conflict-free semantics requires that no conflict should be allowed within the set of accepted arguments.
Admissible semantics requires that one accept (or reject) an argument only if they have reason to do so.
Complete semantics, which is a strengthening of admissible semantics, further demands that one cannot label `undecided' to an argument that should be accepted or rejected.

The following \emph{Tolerability Postulate} captures the idea that conflict should be avoided within the set of acceptability arguments in conflict-free semantics.
It states that attacks between acceptability arguments should be tolerable.

\begin{post}[Tolerability, TP]\label{TP}
A fuzzy labeling  satisfies the \emph{Tolerability Postulate}  over an FAS $\langle\mathcal{A},\mathcal{R}\rangle$ iff  $\forall A\in Args$, $$\max_{B\in Att(A)} B^{a}*\mathcal{R}_{_{BA}}+A^{a}\leq 1.$$
We stipulate that $\max_{B\in Att(A)} B^{a}*\mathcal{R}_{_{BA}}=0$ if $Att(A)=\varnothing$. 
\end{post}

Here we introduce two postulates to refine admissible semantics, which requires providing a reason why an argument is accepted (or rejected) to a certain degree.

The \emph{Weakened Postulate} states that an argument can be rejected to some degree only if it receives the same attack intensity from its acceptability attackers. This postulate extends the classical idea that an argument is labeled as `rejected' only if it has an `accepted' attacker.

\begin{post}[Weakened, WP]\label{WP}
A fuzzy labeling satisfies the \emph{Weakened Postulate}  over an FAS $\langle\mathcal{A},\mathcal{R}\rangle$ iff  $\forall A\in Args$, $A^{r}\leq \max\limits_{B\in Att(A)} B^{a}*\mathcal{R}_{_{BA}}$.
\end{post}

The \emph{Defense Postulate} states that an argument is accepted to some degree only if all of its sufficient attackers are rejected to that degree so that it can be defended to that degree. This postulate extends the classical idea that an argument is labeled as `accepted' only if all of its attackers are labeled as `rejected'. 

\begin{post}[Defense, DP]\label{DP}
A fuzzy labeling satisfies the \emph{Defense Postulate}  over an FAS $\langle\mathcal{A},\mathcal{R}\rangle$ iff  $\forall A\in Args$, $$A^{a}\leq \min_{B\in Att(A)}\{\max \{B^{r},1-\mathcal{A}(B)*\mathcal{R}_{_{BA}}\}\}.$$
We stipulate that $\min_{B\in Att(A)}\{\max \{B^{r},1-\mathcal{A}(B)*\mathcal{R}_{_{BA}}\}\}=1$ if $Att(A)=\varnothing$.
\end{post}

Theorem \ref{Theorem 1} provides an explanation of DP, demonstrating that if a fuzzy labeling satisfies DP, then the acceptability degree of an argument should not be greater than the rejectability degree of its sufficient attackers.

\begin{thm}\label{Theorem 1}
A fuzzy labeling $F\!Lab$ satisfies DP iff for any argument $B\in Att(A)$,  $(B, \mathcal{A}(B))$ sufficiently attacks $(A,A^a)$ implies $A^a\leq B^r$.
\end{thm}

Finally, we establish the postulates to refine complete semantics.
While admissible semantics requires providing a reason for accepting and
rejecting an argument to a certain degree, complete semantics goes further and requires that one cannot leave the degree undecided that should be accepted or rejected.

The \emph{Strict Weakened Postulate} is a strict version of WP.
It states that the rejectability degree of an argument should be equal to the attack intensity of its acceptability attackers.
It ensures that we cannot leave the degree undecided that should be rejected.

\begin{post}[Strict Weakened, SWP]\label{SWP}
A fuzzy labeling satisfies the \emph{Strict Weakened Postulate}  over an FAS $\langle\mathcal{A},\mathcal{R}\rangle$ iff  $\forall A\in Args$, $A^{r}= \max\limits_{B\in Att(A)} B^{a}*\mathcal{R}_{_{BA}}$.
\end{post}

The \emph{Strict Defense Postulate} is the strict version of DP.
It ensures that we cannot leave the degree undecided that should be accepted.

\begin{post}[Strict Defense, SDP]\label{SDP}
A fuzzy labeling satisfies the \emph{Strict Defense Postulate}  over an FAS $\langle\mathcal{A},\mathcal{R}\rangle$ iff  $\forall A\in Args$, 
\begin{displaymath}
  A^{a}=\min\{\min_{B\in Att(A)}\{\max \{B^{r},1-\mathcal{A}(B)*\mathcal{R}_{_{BA}}\}\},\mathcal{A}(A)\}.
\end{displaymath}
\end{post}

Theorem \ref{Theorem 2} provides an explanation of SDP, demonstrating that if a fuzzy labeling satisfies SDP,
then the acceptability degree of an argument is either equal to the lower bound of the rejectability degree of its sufficient attackers or constrained by non-sufficient attackers.

\begin{thm}\label{Theorem 2}
If a fuzzy labeling $F\!Lab$ satisfies SDP, then for any $A\in Args$, $$A^{a}=\min\{\min_{B\in S}B^r,1-\max_{B\notin S}\mathcal{A}(B)*\mathcal{R}_{_{BA}},\mathcal{A}(A)\}$$
 where $S=\{B\in Args\mid (B,\mathcal{A}(B)) \text{ sufficiently attacks } (A,A^a)\}$.
 We stipulate that $\min_{B\in Att(A)}B^{r}=1$ if $Att(A)=\varnothing$.
\end{thm}

Theorem \ref{Theorem 3} shows the link between the above postulates.

\begin{thm}\label{Theorem 3}
The postulates have the following properties:
\begin{enumerate}
\setlength{\itemsep}{0pt}
\setlength{\parsep}{0pt}
\setlength{\parskip}{0pt}
\item $BP+UP+DP$ implies $TP$;
\item $SW\!P$ implies $W\!P$;
\item $SDP$ implies $BP$;
\item $SDP$ implies $DP$.
\end{enumerate}
\end{thm}

\subsection{Fuzzy Labeling Semantics}\label{section Fuzzy Labeling Semantics}

In this section, we will apply fuzzy labeling to generalize classical semantics.
Roughly, fuzzy labeling semantics can be regarded as a quantitative generalization of classical labeling semantics.
In all the following definitions, we consider a fixed FAS $\mathcal{F}=\langle\mathcal{A},\mathcal{R}\rangle$ and a fuzzy labeling $F\! Lab$.

We start by introducing \emph{conflict-free fuzzy labeling}.
Conflict-free is always a primary requirement: no conflict should be allowed within the set of acceptability arguments. 
The corresponding postulates are BP, UP and TP.

\begin{defn}[Conflict-free Fuzzy Labeling]
\label{conflict-free}
A fuzzy labeling is \emph{conflict-free} iff it satisfies BP, UP and TP.
\end{defn}

We now define \emph{admissible fuzzy labeling}, which requires that for every argument one accepts (or rejects) to some degree, a reason why it is accepted (or rejected) to that degree should be provided. 
The corresponding postulates are BP, UP, WP and DP.

\begin{defn}[Admissible Fuzzy Labeling]\label{admissible}
A fuzzy labeling is \emph{admissible} iff it satisfies BP, UP, WP and DP.
\end{defn}

While admissible fuzzy labeling requires providing a reason for accepting and rejecting an argument to a certain degree, \emph{complete fuzzy labeling} goes further and requires that one should not leave the degree undecided that should be accepted or rejected.
The corresponding postulates are BP, UP, SWP and SDP.

\begin{defn}[Complete Fuzzy Labeling]
\label{complete}
A fuzzy labeling is \emph{complete} iff it satisfies BP, UP, SWP and SDP.
\end{defn}

Lemma \ref{lemma1} implies that a complete fuzzy labeling is uniquely defined by the set of acceptability arguments or the set of rejectability arguments.
\begin{lem}\label{lemma1}
Let $F\! Lab_{1}$ and $F\! Lab_{2}$ be two complete fuzzy labelings of an FAS $\langle\mathcal{A},\mathcal{R}\rangle$.
It holds that
$F\! Lab_{1}^a\subseteq F\! Lab_{2}^a \text{ iff } F\! Lab_{1}^r\subseteq F\! Lab_{2}^r$.
\end{lem}

In the following, we refine several widely studied classical semantics, including grounded, preferred, semi-stable and stable, by imposing constraints such as maximality or minimality on complete semantics.

We now refine classical grounded semantics, which is characterized by minimal accepted arguments and is generally considered as the least questionable semantics.
\emph{Grounded fuzzy labeling} states that the set of acceptability arguments should be minimal among all complete fuzzy labelings.

\begin{defn}[Grounded Fuzzy Labeling]\label{grounded}
$F\! Lab$ is a \emph{grounded fuzzy labeling} iff it is a complete fuzzy labeling where ${F\! Lab}^a$ is minimal (w.r.t. fuzzy set inclusion) among all complete fuzzy labelings.
\end{defn}

The following proposition can be easily derived from Lemma \ref{lemma1}.
\begin{prop}\label{prop2}
The following statements are equivalent:
\begin{enumerate}
\setlength{\itemsep}{0pt}
\setlength{\parsep}{0pt}
\setlength{\parskip}{0pt}
\item $F\!Lab$ is a complete fuzzy labeling where ${F\! Lab}^a$ is minimal (w.r.t. fuzzy set inclusion) among all complete fuzzy labelings;
\item $F\!Lab$ is a complete fuzzy labeling where ${F\! Lab}^r$ is minimal (w.r.t. fuzzy set inclusion) among all complete fuzzy labelings;
\item $F\!Lab$ is a grounded fuzzy labeling.
\end{enumerate}
\end{prop}

Proposition \ref{unique grounded} states that every FAS has exactly one grounded fuzzy labeling.

\begin{prop}\label{unique grounded}
Every FAS has a unique grounded fuzzy labeling.
\end{prop}

We now refine classical preferred semantics, which is characterized by maximal accepted arguments.
\emph{Preferred fuzzy labeling} states that the set of acceptability arguments should be maximal among all complete fuzzy labelings.

\begin{defn}[Preferred Fuzzy Labeling]\label{preferred}
$F\! Lab$ is a \emph{preferred fuzzy labeling} iff it is a complete fuzzy labeling where ${F\! Lab}^a$ is maximal (w.r.t. fuzzy set inclusion) among all complete fuzzy labelings.
\end{defn}

The following proposition can be easily derived from Lemma \ref{lemma1}.
\begin{prop}\label{Proposition 4}
The following statements are equivalent:
\begin{enumerate}
\setlength{\itemsep}{0pt}
\setlength{\parsep}{0pt}
\setlength{\parskip}{0pt}
\item $F\!Lab$ is a complete fuzzy labeling where ${F\! Lab}^a$ is maximal (w.r.t. fuzzy set inclusion) among all complete fuzzy labelings;
\item $F\!Lab$ is a complete fuzzy labeling where ${F\! Lab}^r$ is maximal (w.r.t. fuzzy set inclusion) among all complete fuzzy labelings;
\item $F\!Lab$ is a preferred fuzzy labeling.
\end{enumerate}
\end{prop}

Next, we refine classical semi-stable semantics, which is characterized by minimal undecided arguments.
\emph{Semi-stable fuzzy labeling} requires that the set of undecidability arguments should be minimal among all complete fuzzy labelings.

\begin{defn}[Semi-stable Fuzzy Labeling]\label{semi-stable}
$F\! Lab$ is a \emph{semi-stable fuzzy labeling} iff it is a complete fuzzy labeling where ${F\! Lab}^u$ is minimal (w.r.t. fuzzy set inclusion) among all complete fuzzy labelings.
\end{defn}

Finally, we refine classical stable semantics, which is characterized by the forbidden of undecided arguments.
\emph{Stable fuzzy labeling} requires that the set of undecidability arguments should be empty.

\begin{defn}[Stable Fuzzy Labeling]\label{stable}
$F\! Lab$ is a \emph{stable fuzzy labeling} iff it is a complete fuzzy labeling with $F\! Lab^u=\varnothing$.
\end{defn}

Theorem \ref{semantics inclusions} shows the relations between fuzzy labeling semantics. 

\begin{thm}[semantics inclusions]\label{semantics inclusions}
Let $\mathcal{F}=\langle\mathcal{A},\mathcal{R}\rangle$ be a fuzzy argumentation system  and $F\! Lab$ be a fuzzy labeling.
It holds that
\begin{enumerate}
\setlength{\itemsep}{0pt}
\setlength{\parsep}{0pt}
\setlength{\parskip}{0pt}

  \item  if $F\! Lab$ is admissible then it is conflict-free;
  \item  if $F\! Lab$ is complete then it is admissible;
  \item  if $F\! Lab$ is grounded/preferred then it is complete;
  \item  if $F\! Lab$ is semi-stable then it is preferred;
  \item  if $F\! Lab$ is stable then it is semi-stable.
\end{enumerate}
\end{thm}

According to Theorem \ref{semantics inclusions}, the relations between fuzzy labeling semantics are identical to that of classical labeling semantics. Next, we illustrate these fuzzy labeling semantics with the following two examples.

\begin{exmp}\label{conflict-freeeg}
Consider a fuzzy argumentation system over $Args=\{A, B, C\}$\begin{displaymath}
\mathcal{F}=\langle\{(A,0.8),(B,0.7),(C,0.6)\},\{\big( (A,B),1\big), \big( (B,C),0.9\big)\}\rangle.
\end{displaymath}
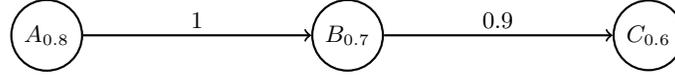
\begin{figure}[H]
\centering
\begin{tikzpicture}[
roundnode/.style={circle, draw=black!100, fill=white!5, thick, minimum size=5mm},
rectanglenode/.style={rectangle, draw=black!0, fill=white!5, thick, minimum size=0mm},
]
\node[roundnode]      (1) at(-2,0)      {$A_{0.8}$};
\node[roundnode]      (2) at(2,0)      {$B_{0.7}$};
\node[roundnode]      (3) at(6,0)      {$C_{0.6}$};
\draw[->,thick] (1)--(2);
\draw[->,thick] (2)--(3);
\node at(0,0.2) {1};
\node at(4,0.2) {0.9};
\end{tikzpicture}

\caption{The Graph Representation of $\mathcal{F}$}
\end{figure}

Consider two fuzzy labelings $F\!Lab_1$ and $F\!Lab_2$.

$F\!Lab_1$ is given as

\begin{itemize}
\setlength{\itemsep}{0pt}
\setlength{\parsep}{0pt}
\setlength{\parskip}{0pt}
\item[] $A^{a_1}=0.5, A^{r_1}=0,  A^{u_1}=0.5;$
\item[] $B^{a_1}=0.4, B^{r_1}=0.5, B^{u_1}=0.1;$
\item[] $C^{a_1}=0.6, C^{r_1}=0.4, C^{u_1}=0$.
\end{itemize}

$F\!Lab_2$ is given as 
\begin{itemize}
\setlength{\itemsep}{0pt}
\setlength{\parsep}{0pt}
\setlength{\parskip}{0pt}
 \item[] $A^{a_2}=0.8, A^{r_2}=0, A^{u_2}=0.2;$
 \item[] $B^{a_2}=0.2, B^{r_2}=0.8, B^{u_2}=0;$ 
 \item[] $C^{a_2}=0.6, C^{r_2}=0.2, C^{u_2}=0.2$.
\end{itemize}

It is evident that both $F\!Lab_2$ and $F\!Lab_2$ satisfy BP and UP.
Since
\begin{align*}
  A^{a_1}*\mathcal{R}_{_{AB}}+B^{a_1}&=\min\{0.5,1\}+0.4\leq 1,\\
  B^{a_1}*\mathcal{R}_{_{BC}}+C^{a_1}&=\min\{0.4,0.9\}+0.6\leq 1,
\end{align*}
it follows that $F\! Lab_{1}^a$ satisfies TP.
Therefore, $F\!Lab_{1}$ is conflict-free.
And the equations
\begin{align*}
  B^{r_1}&=A^{a_1}*\mathcal{R}_{_{AB}}=\min\{0.5,1\}\leq 0.5,\\
  C^{r_1}&=B^{a_1}*\mathcal{R}_{_{BC}}=\min\{0.4,0.9\}\leq 0.4,
\end{align*}
ensure that $F\! Lab_{1}^a$ satisfies WP.
However, since $(B,0.7)$ sufficiently attacks $(C,0.6)$ but $C^{a_1}> B^{r_1}$, it follows that $F\!Lab_{1}$ violates DP by Theorem \ref{Theorem 1} and thus it is not admissible.

As for $F\!Lab_2$, we check that $F\!Lab_{2}$ satisfies SDP and SWP.
\begin{align*}
  A^{a_2}&=\min\{\min_{X\in Att(A)}\{\max \{X^{r},1-\mathcal{A}(X)*\mathcal{R}_{_{XA}}\}\},\mathcal{A}(A)\}\\
 &=\min\{1,\mathcal{A}(A)\}= 0.8\\
  B^{a_2}&=\min\{\min_{X\in Att(B)}\{\max \{X^{r},1-\mathcal{A}(X)*\mathcal{R}_{_{XB}}\}\},\mathcal{A}(B)\}\\
 &=\min\{1-\mathcal{A}(A)*\mathcal{R}_{_{AB}},\mathcal{A}(B)\}=0.2\\
    \end{align*}
  \begin{align*}
  C^{a_2} &=\min\{\min_{X\in Att(C)}\{\max \{X^{r},1-\mathcal{A}(X)*\mathcal{R}_{_{XC}}\}\},\mathcal{A}(C)\}\\
&=\min\{\max\{B^{r_2},1-\mathcal{A}(B)*\mathcal{R}_{_{BC}}\},\mathcal{A}(C)\}= 0.6\\
  A^{r_2} &=\max_{X\in Att(A)} B^{a}*\mathcal{R}_{_{XB}}=0\\
 B^{r_2} &=\max_{X\in Att(B)} B^{a}*\mathcal{R}_{_{XB}}=A^{a_2}*\mathcal{R}_{_{AB}}=0.8\\
 C^{r_2} &=\max_{X\in Att(C)} B^{a}*\mathcal{R}_{_{XC}}=B^{a_2}*\mathcal{R}_{_{BC}}=0.2
\end{align*}
Therefore $F\!Lab_{2}$ is admissible and also complete.
\end{exmp}

\begin{exmp}\label{circleexample}
Consider a fuzzy argumentation system with a cycle

$$\mathcal{F}=\langle\{(A, 0.8), (B, 0.6)\}, \{\big( (A, B),1\big), \big( (B, A), 1\big)\}\rangle.$$

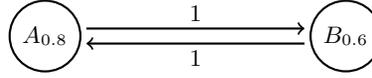
\begin{figure}[H]
\centering
\begin{tikzpicture}[
roundnode/.style={circle, draw=black!100, fill=white!5, thick, minimum size=5mm},
rectanglenode/.style={rectangle, draw=black!0, fill=white!5, thick, minimum size=0mm},
]
\node[roundnode]      (1) at(-2,0)      {$A_{0.8}$};
\node[roundnode]      (2) at(2,0)      {$B_{0.6}$};
\draw[->,thick] (-1.45,0.1)--(1.45,0.1);
\draw[->,thick] (1.45,-0.1)--(-1.45,-0.1);
\node at(0,0.3) {1};
\node at(0,-0.3) {1};
\end{tikzpicture}
\caption{The Graph Representation of $\mathcal{F}$}
\end{figure}

Consider three fuzzy labelings $F\!Lab_1$, $F\!Lab_2$ and $F\!Lab_3$, where

$F\!Lab_1(A)=(0.8,0.2,0)$, $F\!Lab_1(B)=(0.2,0.8,0)$;

$F\!Lab_2(A)=(0.4,0.6,0)$, $F\!Lab_2(B)=(0.6,0.4,0)$; 

$F\!Lab_3(A)=(0.4,0.2,0.4)$, $F\!Lab_3(B)=(0.2,0.4,0.4)$.

According to Definition \ref{complete}, all of these fuzzy labelings are complete.
Since $F\! Lab_{1}^a$ and $F\! Lab_{2}^a$ are maximal among all complete fuzzy labelings, it follows that $F\! Lab_{1}$ and $F\! Lab_{2}$ are preferred.
Analogously, it is clear that $F\! Lab_{3}^a$ is minimal among all complete fuzzy labelings, and thus $F\! Lab_{3}$ is grounded.
Since $F\! Lab_{1}^u=F\! Lab_{2}^u=\varnothing$,   $F\! Lab_{1}$ and $F\! Lab_{2}$ are both semi-stable and stable.
\end{exmp}

\section{Comparison to Related Work}\label{section relationship}
\subsection{Relation to Fuzzy Extension Semantics}\label{relation to GFAS}

In this section, we examine the relationship between fuzzy labeling semantics and fuzzy extension semantics.
The fuzzy extension semantics introduced in \cite{wu2016godel} are listed as follows.

\begin{defn}[\cite{wu2016godel}] \label{extension of FAS}
Let $\mathcal{F}=\langle\mathcal{A}, \mathcal{R}\rangle$ be an FAS and $E\subseteq \mathcal{A}$ be a fuzzy set.

\textbf{Weakening Defense}: If there is a sufficient attack relation $\big( (A,B), \mathcal{R}_{_{AB}}\big)$ from $(A, a)$ to $(B, b)$, then we say that $(A,a)$ \emph{weakens} $(B,b)$ to $(B,b')$, where $b'=1-a*\mathcal{R}_{_{AB}}.$
A fuzzy set $S\subseteq\mathcal{A}$ \emph{weakening defends} a fuzzy argument $(C, c)$ in $\mathcal{A}$ if, for any $B\in Att(A)$, if $(B, \mathcal{A}(B))$ sufficiently attacks $(C,c)$, then there is some $(A, a) \in S$ such that $(A, a)$ weakens $(B, \mathcal{A}(B))$ to $(B, b^{'})$ and $(B, b^{'})$ tolerably attacks $(C, c)$.

The fuzzy extension semantics are defined over fuzzy set. 
\begin{itemize}
\setlength{\itemsep}{0pt}
\setlength{\parsep}{0pt}
\setlength{\parskip}{0pt}
    \item A fuzzy set $E$ is a \emph{conflict-free} fuzzy extension if all attacks in $E$ are tolerable.
    \item A conflict-free fuzzy extension $E$ is \emph{admissible} if $E$ weakening defends each element in $E$.
    \item An admissible fuzzy extension $E$ is \emph{complete} if it contains all the fuzzy arguments in $\mathcal{A}$ that $E$ weakening defends.
    \item The \emph{grounded} fuzzy extension is the minimal  complete fuzzy extension.
    \item A \emph{preferred} fuzzy extension is a maximal  complete fuzzy extension.
\end{itemize}
\end{defn}

Given the fuzzy labeling semantics of FAS, we can establish a correspondence relationship with fuzzy extension semantics through the mapping functions $Ext2F\!Lab$ and $F\!Lab2Ext$.
Roughly speaking, the set of acceptability arguments can be regarded as a fuzzy extension through the transform functions.

\begin{defn}
\label{the translation FLab2Ext}
Given an FAS $\mathcal{F}=\langle\mathcal{A}, \mathcal{R}\rangle$ and a fuzzy labeling $F\! Lab$, the corresponding fuzzy extension $F\!Lab2Ext$ is defined as $F\!Lab2Ext(F\! Lab)={F\! Lab}^a$.
\end{defn}

\begin{defn}
\label{the translation Ext2FLab}
Given an FAS $\mathcal{F}=\langle\mathcal{A}, \mathcal{R}\rangle$ and a fuzzy extension $E$, the corresponding fuzzy labeling $Ext2F\!Lab(E)$ is defined as $Ext2F\!Lab(E)=\{E, E^{+}, (E\oplus E^{+})^c\}$ where
\begin{align*}
    E^{+}&=\{(A, a_{d})\mid A\in Args\text{ and }a_{d}=\max\limits_{B \in Args}E(B)*\mathcal{R}_{_{BA}}\}\\
(E\oplus E^{+})^c&=\{(A, a_{d})\mid A\in Args \text{ and } a_{d}=1-E(A)-E^+(A)\}.
\end{align*}
\end{defn}

The following theorem examines the relations between fuzzy labeling semantics and fuzzy extension semantics.

\begin{thm}\label{correspondence relation for cac}
Let $\mathcal{F}=\langle\mathcal{A}, \mathcal{R}\rangle$ be a fuzzy argumentation system.
For a semantics $\mathcal{S}\in$ \{conflict-free, admissible, complete, grounded, preferred\}, 
it holds that:
\begin{enumerate}
\setlength{\itemsep}{0pt}
\setlength{\parsep}{0pt}
\setlength{\parskip}{0pt}
\item if $E$ is an $\mathcal{S}$ fuzzy extension, then $Ext2F\!Lab(E)$ is an $\mathcal{S}$ fuzzy labeling;
\item if $F\! Lab$ is an $\mathcal{S}$ fuzzy labeling, then $F\!Lab2Ext(F\! Lab)$ is an $\mathcal{S}$ fuzzy extension.
\end{enumerate}
\end{thm}

The following theorem states that complete fuzzy labelings and complete fuzzy extensions stand in a one-to-one correspondence relationship with each other,
and this relationship also holds for grounded and preferred semantics.

\begin{thm}\label{Theorem 11}
For any FAS $\mathcal{F}=\langle\mathcal{A}, \mathcal{R}\rangle$, $F\!Lab$ is a complete  (resp. grounded, preferred) fuzzy labeling iff there is a complete  (resp. grounded, preferred) fuzzy extension $E$ s.t. $F\!Lab=Ext2F\!Lab(E)$.
\end{thm}

For semantics in \{conflict-free, admissible, complete, grounded, preferred\}, the fuzzy extension semantics can be regarded as a special version of the fuzzy labeling semantics.
The latter is a more general approach to character semantics which provide a clearer status for arguments. 
This correspondence  is similar to that of the classical labeling semantics and classical extension semantics.

\subsection{Relation to Classical Labeling Semantics}

In this section, we examine the relationship between fuzzy labeling semantics and classical labeling semantics.
Let us provide the notions of argumentation framework \cite{dung1995acceptability} and classical labeling semantics \cite{baroni2011introduction,caminada2006issue}.

\begin{defn}[\cite{dung1995acceptability}]
An \emph{argumentation framework} (AF) is a pair $(Args,Att)$ where $Args$ is a set of \emph{arguments} and $Att\subseteq Args\times Args$ is a set of \emph{attacks}.
An argument $A$ \emph{attacks} an argument $B$ iff $(A,B)\in Att$, and $A$ is called the \emph{attacker} of $B$. 
\end{defn}

\begin{defn}[\cite{baroni2011introduction,caminada2006issue}]\label{argumentlabeling}
Let $AF=(Args, Att)$ be an argumentation framework. An argument labeling is a total function $Lab: Args \rightarrow \{in, out, undec\}$ where $in$, $out$, and $undec$ represent accepted, rejected, and undecided respectively.
An argument labeling $Lab$ is usually represented as a triple $(in(Lab), out(Lab), undec(Lab))$ where $in(Lab)=\{A\in Args\vert Lab(A)=in\}$, $out(Lab)=\{A\in Args\vert Lab(A)=out\}$, and $undec(Lab)=\{A\in Args\vert Lab(A)=undec\}$.

$Lab$ is a \emph{conflict-free} labeling iff for each argument $A\in in(Lab)$, there exists no argument $B\in Att(A)$ s.t. $Lab(B)=in$.

$Lab$ is an \emph{admissible} labeling iff for each argument $A\in Args$ it holds that:
 \begin{enumerate}
\setlength{\itemsep}{0pt}
\setlength{\parsep}{0pt}
\setlength{\parskip}{0pt}
\item if $A$ is labelled $in$, then all its attackers are labelled out;
\item if $A$ is labelled $out$, then it has at least one attacker that is labelled $in$.
\end{enumerate}

$Lab$ is a \emph{complete} labeling iff for each argument $A\in Args$ it holds that:
 \begin{enumerate}
\setlength{\itemsep}{0pt}
\setlength{\parsep}{0pt}
\setlength{\parskip}{0pt}
\item if $A$ is labelled $in$, then all its attackers are labelled out;
\item if $A$ is labelled $out$, then it has at least one attacker that is labelled $in$.
\item if $A$ is labelled $undec$, then not all its attackers are labelled $out$ and it does not have an attacker that is labelled $in$.
\end{enumerate}

The \emph{grounded} labeling is a complete labeling $Lab$ where $in(Lab)$ is minimal (w.r.t. set inclusion) among all complete labelings.

A \emph{preferred} labeling is a complete labeling $Lab$ where $in(Lab)$ is maximal (w.r.t. set inclusion) among all complete labelings.

A \emph{semi-stable} labeling is a complete labeling $Lab$ where $undec(Lab)$ is minimal (w.r.t. set inclusion) among all complete labelings.

A \emph{stable} labeling is a complete labeling $Lab$ where $undec(Lab)=\varnothing$.
\end{defn}
 
We provide a transformation of AF to FAS and subsequently adapt classical argument labeling to fuzzy labeling.

\begin{defn}\label{correspond}
Given an argumentation framework $AF=(Args,Att)$, the corresponding FAS $\langle\mathcal{A},\mathcal{R}\rangle$ is defined as follows:
\begin{itemize}
\setlength{\itemsep}{0pt}
\setlength{\parsep}{0pt}
\setlength{\parskip}{0pt}
    \item if $A\in Args$, then $\mathcal{A}(A)=1$;
    \item if $A\notin Args$, then $\mathcal{A}(A)=0$;
    \item if $(A,B)\in Att$, then $\mathcal{R}(A,B)=1$;
    \item if $(A,B)\notin Att$, then $\mathcal{R}(A,B)=0$.
\end{itemize}
Given a classical argument labeling $Lab$,
the corresponding fuzzy labeling $F\! Lab$ is defined as follows:
\begin{itemize}
\setlength{\itemsep}{0pt}
\setlength{\parsep}{0pt}
\setlength{\parskip}{0pt}
    \item if $Lab(A)=in$, then $F\! Lab(A)=(1,0,0)$;
    \item if $Lab(A)=out$, then $F\! Lab(A)=(0,1,0)$;
    \item if $Lab(A)=undec$, then $F\! Lab(A)=(0,0,1)$.
\end{itemize}
\end{defn}

The following theorem shows the relationship between fuzzy labeling semantics and classical labeling semantics.

\begin{thm}\label{Theorem 12}
Let $AF=(Args, Att)$ be an argumentation framework and  $Lab$ be an argument labeling of $AF$.
For a semantics $\mathcal{S}\in $\{conflict-free, admissible, complete, grounded, stable\}, 
if $Lab$ is an $\mathcal{S}$ labeling of $AF$, then the corresponding fuzzy labeling is also an $\mathcal{S}$ fuzzy labeling of the corresponding FAS.
\end{thm}

Theorem \ref{Theorem 12} shows that for a semantics $\mathcal{S}\in\{$conflict-free, admissible, complete, grounded, stable\}, each $\mathcal{S}$ labeling of AF is an $\mathcal{S}$ fuzzy labeling of the corresponding FAS.
The results prove that fuzzy labeling semantics are compatible with classical labeling semantics, especially for grounded semantics which is unique.
Unfortunately, this relationship does not hold for preferred and semi-stable semantics when dealing with AFs containing odd cycles.
Considering an AF with a self-attacking argument $(\{A\},\{(A,A)\})$, the unique complete labeling is $(\varnothing,\varnothing,\{A\})$.
However, in the corresponding FAS, $(\{(A,0.5)\},\{(A,0.5)\},\emptyset)$ is a preferred fuzzy labeling where $\{(A,0.5)\}$ is not empty.

\subsection{Other Related Work}\label{Discussion}

In this section, we discuss related work on the evaluation of arguments in various QuAS.

Many researchers focused on the semantics that consider the acceptability degree alone in QuAS.
In \cite{da2011changing}, da Costa Pereira et al. introduced trust-based semantics for FAS.
In \cite{gabbay2015equilibrium}, Gabbay and Rodrigues introduced Iterative-based semantics for numerical AF.
In \cite{Amgoud2017Acceptability,Amgoud2021Evaluation}, Amgoud et al. proposed weighted max-based, card-based and h-categorizer semantics for WAS.
In \cite{amgoud2016evaluation,Amgoud2018Evaluation}, Amgoud and Ben-Naim proposed top-based, reward-based, aggregation-based and exponent-based semantics for WAS with support relation.
In \cite{baroni2015automatic}, Baroni et al. proposed the QuAD semantics for acyclic Quantitative Bipolar AF, which was later extended to DF-QuAD semantics by Rago et al. \cite{rago2016discontinuity}.
These works aim to compute the acceptability degree alone in the context of QuAS.

Extension or labeling semantics for QuAS have also been studied in the literature. 
In \cite{wu2016godel}, Wu et al. proposed fuzzy extension semantics over fuzzy set for FAS, such as grounded, preferred, etc.
In \cite{janssen2008fuzzy}, Janssen et al. proposed extension semantics for FAS, such as x-stable, y-preferred, etc.
In \cite{Bistarelli2018novel}, Bistarelli et al. redefined extension semantics for WAS by considering weighted defence.
In \cite{Stefano2022labeling}, Bistarelli and Taticchi redefined labeling semantics for WAS by assigning each argument a label in $\{in, out, undec\}$.
In \cite{dunne2011weighted}, Dunne et al. obtained the extension semantics of WAS by disregarding the attacks whose total weight is less than a given budget.

Obviously, our fuzzy labeling semantics differs from the evaluation methods in abstract argumentation, as it provides a richer scale for argument strength by associating each argument with degrees of acceptability/rejectability/undecidability.
It is worth noting that this type of evaluation methodology is widely employed in many areas.
For instance, in Dempster-Shafer theory, each assertion is associated with three non-negative degrees $(p,q,r)$ s.t. $p+q+r=1$. Here, $p$ is the probability ``for'' the assertion, $q$ is the probability ``against'' the assertion, and $r$ is the probability of ``don't know'' \cite{dempster2008dempster,shafer1976mathematical}.
Similarly in \cite{josang1997artificial}, an agent's opinion is associated with a triple $(b,d,i)$, where $b$ for the degree of `belief', $d$ for the degree of `disbelief', and $i$ for the degree of `ignorance' in the field of subjective logic.
In \cite{haenni2009probabilistic}, Haenni evaluated arguments (not in Dung-style argumentation) using the \emph{degrees of belief/disbelief/ignorance}, and discussed the desirable properties of this method, particularly the \emph{non-additivity} for classifying disbelief and ignorance.



\section{Conclusion and Future Work}\label{conclusion}

In this paper, we proposed a more comprehensive evaluation method called fuzzy labeling for fuzzy argumentation systems, which  describes the argument strength as a triple consisting of acceptability, rejectability, and undecidability degrees.
Such a  setting  sheds new light on defining argument strength and provides a deeper understanding of the status of arguments.
For the purpose of evaluating arguments, we provided a class of fuzzy labeling semantics which generalize the classical semantics, such as complete, semi-stable, etc.
Finally, we examined the relationships between fuzzy labeling semantics and existing semantics in the literature.

The fuzzy labeling theory provides a new way to evaluate argument strength. 
This work can be extended in several directions. 
Firstly, it is important to study the properties of fuzzy labeling semantics. 
Secondly, there is ample room for exploring fuzzy labeling semantics for QuAS, especially utilizing the recently growing single-status approach \cite{Amgoud2021Evaluation}.
Finally, it would be interesting to develop fuzzy labeling applications for decision systems, judgment aggregation, algorithms, and other related fields.



\bibliographystyle{splncs04}
\bibliography{mybibliography}

\appendix
\section*{Appendix}
\setcounter{equation}{0}

\renewcommand\theequation{\arabic{equation}} 
This appendix contains the proofs of the results presented in the paper.

\noindent \textbf{Proof of Theorem \ref{Theorem 1}:}

First, we prove the forward direction. 
Let  $F\!Lab$ be a fuzzy labeling of FAS $\langle\mathcal{A},\mathcal{R}\rangle$ and it satisfies DP. Then for any argument $A\in Args$, 
$$A^a\leq \min_{B\in Att(A)}\{\max \{B^{r},1-\mathcal{A}(B)*\mathcal{R}_{_{BA}}\}\}.$$
For any argument $B\in Att(A)$, if $(B, \mathcal{A}(B))$ sufficiently attacks $(A, A^a)$, then $1-\mathcal{A}(B)*\mathcal{R}_{_{BA}}<A^a$.
  Therefore, $A^a\leq \max \{B^{r},1-\mathcal{A}(B)*\mathcal{R}_{_{BA}}\}=B^{r}$.
  
Next we prove the reverse direction. Suppose for any argument $B\in Att(A)$, if $(B, \mathcal{A}(B))$ sufficiently attacks $(A,A^a)$, then $A^a\leq B^r$.
It directly follows that 
$$A^a\leq \min_{B\in S}\{\max\{B^r, 1-\mathcal{A}(B)*\mathcal{R}_{_{BA}}\}\}$$
 where $S=\{B\in Args\mid (B,\mathcal{A}(B)) \text{ sufficiently attacks } (A,A^a)\}$.
And for any argument $B\notin S$, it is obvious that 
\begin{align*}
A^a &\leq \min_{B\notin S}\{1-\mathcal{A}(B)*\mathcal{R}_{_{BA}}\}\\
 &\leq \min_{B\notin S}\{\max\{B^r, 1-\mathcal{A}(B)*\mathcal{R}_{_{BA}}\}\}.
\end{align*}
Therefore, $A^{a}\leq \min_{B\in Att(A)}\{\max \{B^{r},1-\mathcal{A}(B)*\mathcal{R}_{_{BA}}\}\}$.

\noindent \textbf{Proof of Theorem \ref{Theorem 2}:}

Let  $F\!Lab$ be a fuzzy labeling of FAS $\langle\mathcal{A},\mathcal{R}\rangle$ and it satisfies SDP. 
Then for any argument $A\in Args$, $$A^{a}= \min\{\min_{B\in Att(A)}\{\max \{B^{r},1-\mathcal{A}(B)*\mathcal{R}_{_{BA}}\}\},\mathcal{A}(A)\}.$$
Let $S=\{B\in Args\mid (B,\mathcal{A}(B)) \text{ sufficiently attacks } (A,A^a)\}$.
Then it follows from Theorem \ref{Theorem 1} that $A^a\leq \min_{B\in S}B^r$,
and it is obvious that 
$A^a\leq 1-\max_{B\notin S}\mathcal{A}(B)*\mathcal{R}_{_{BA}}$.
Thus
\begin{equation*}
A^a\leq \min\{\min_{B\in S}B^r,1-\max_{B\notin S}\mathcal{A}(B)*\mathcal{R}_{_{BA}},\mathcal{A}(A)\}.
\end{equation*}
Now we need to prove that $A^a\geq \min\{\min_{B\in S}B^r,1-\max_{B\notin S}\mathcal{A}(B)*\mathcal{R}_{_{BA}},\mathcal{A}(A)\}$.
There are two cases.
If 
\begin{align*}
      \min_{B\in S}\{\max \{B^{r},1-\mathcal{A}(B)*\mathcal{R}_{_{BA}}\}\}
      \geq \min_{B\notin S}\{\max\{B^{r},1-\mathcal{A}(B)*\mathcal{R}_{_{BA}}\}\},
  \end{align*} 
    then \begin{align*}
A^{a}=&\min\{\min_{B\in Att(A)}\{\max \{B^{r},1-\mathcal{A}(B)*\mathcal{R}_{_{BA}}\}\},\mathcal{A}(A)\}\\
=&\min\{\min_{B\notin S}\{\max \{B^{r},1-\mathcal{A}(B)*\mathcal{R}_{_{BA}}\}\},\mathcal{A}(A)\}\\
\geq& \min\{1-\max_{B\notin S}\mathcal{A}(B)*\mathcal{R}_{_{BA}},\mathcal{A}(A)\}\\
\geq& \min\{\min_{B\in S}B^r,1-\max_{B\notin S}\mathcal{A}(B)*\mathcal{R}_{_{BA}},\mathcal{A}(A)\}\\
\geq & A^a.
\end{align*}
Therefore,
\begin{align*}
    A^a
    = &\min\{\min_{B\in S}B^r,1-\max_{B\notin S}\mathcal{A}(B)*\mathcal{R}_{_{BA}},\mathcal{A}(A)\}.
\end{align*}
  And if \begin{align*}
      \min_{B\in S}\{\max \{B^{r},1-\mathcal{A}(B)*\mathcal{R}_{_{BA}}\}\}
      \leq \min_{B\notin S}\{\max\{B^{r},1-\mathcal{A}(B)*\mathcal{R}_{_{BA}}\}\},
  \end{align*} 
  then, 
    \begin{align*}
    A^a &= \min\{\min_{B\in Att(A)}\{\max \{B^{r},1-\mathcal{A}(B)*\mathcal{R}_{_{BA}}\}\},\mathcal{A}(A)\}\\
    &=\min\{\min_{B\in S}\{\max \{B^{r},1-\mathcal{A}(B)*\mathcal{R}_{_{BA}}\}\},\mathcal{A}(A)\}\\
    &\geq \min\{\min_{B\in S}B^r,\mathcal{A}(A)\}\\
    &\geq \min\{\min_{B\in S}B^r,1-\max_{B\notin S}\mathcal{A}(B)*\mathcal{R}_{_{BA}},\mathcal{A}(A)\}\\
    &\geq A^a.
  \end{align*}
Consequently, combining the two cases, $$A^a=\min\{\min_{B\in S}B^r,1-\max_{B\notin S}\mathcal{A}(B)*\mathcal{R}_{_{BA}},\mathcal{A}(A)\}.$$

\noindent \textbf{Proof of Theorem \ref{Theorem 3}:}

1. Let $F\!Lab$ be a fuzzy labeling of FAS $\langle\mathcal{A},\mathcal{R}\rangle$. 
Assume that $F\!Lab$ satisfies BP,UP, and DP.
If it does not satisfy TP, then there is an argument $A\in Args$ such that $A^{a}> 1-C^{a}*\mathcal{R}_{_{CA}}$ holds for some $C\in Att(A)$.
We then have
\begin{align*}
   A^{a} \leq & \min_{B\in Att(A)}\{\max \{B^{r},1-\mathcal{A}(B)*\mathcal{R}_{_{BA}}\}\} &(\mbox{By DP})\\
    \leq & \max \{C^{r},1-\mathcal{A}(C)*\mathcal{R}_{_{CA}}\}  \\
    \leq & \max \{C^{r},1-C^{a}*\mathcal{R}_{_{CA}}\} &(\mbox{By  BP})\\
    = & 1-C^{a}*\mathcal{R}_{_{CA}} &(\mbox{By UP})\\
       < & A^{a}. 
  \end{align*}
Thus a contradiction occurs. 

2-4 are Trivial.

\noindent \textbf{Proof of Lemma \ref{lemma1}:}

$\Rightarrow$: Suppose $F\! Lab_{1}^a\subseteq F\! Lab_{2}^a$.
Then $F\! Lab_{1}^a(A)\leq F\! Lab_{2}^a(A)$ holds for any argument $A\in Args$.
Therefore, it follows from SWP that $F\! Lab_{1}^r(A)\leq F\! Lab_{2}^r(A)$ holds for any argument $A\in Args$.
Hence, $F\! Lab_{1}^r\subseteq F\! Lab_{2}^r$.

$\Leftarrow$: Suppose $F\! Lab_{1}^r\subseteq F\! Lab_{2}^r$.
Then $F\! Lab_{1}^r(A)\leq F\! Lab_{2}^r(A)$ holds for any argument $A\in Args$.
Therefore, it follows from SDP that $F\! Lab_{1}^a(A)\leq F\! Lab_{2}^a(A)$ holds for any argument $A\in Args$.
Hence, $F\! Lab_{1}^a\subseteq F\! Lab_{2}^a$.

\noindent \textbf{Proof of Proposition \ref{prop2}:}

The proof is straightforward by Lemma \ref{lemma1}.

\noindent \textbf{Proof of Proposition \ref{unique grounded}:}

According to Definition 10 in \cite{wu2016godel}, the grounded fuzzy extension is the least fixed point of the characteristic function $F$, where $F$ is a function from the set of all the subsets of $\mathcal{A}$ to itself, such that for any $S\subseteq \mathcal{A}$, $F(S)=\{(A,a)\vert$ $S$  weakening defends  $(A,a)\}$.
Since $F$ is monotonically increasing (w.r.t. fuzzy set inclusion), it follows from the Knaster-Tarski theorem that $F$ has a unique smallest fixed point.
Thus, every FAS has a unique grounded fuzzy extension.
From Theorem \ref{Theorem 11}, the grounded fuzzy labeling and the grounded fuzzy extension stand in a one-to-one correspondence relationship.
Therefore, every FAS has a unique grounded fuzzy labeling.

\noindent \textbf{Proof of Proposition \ref{Proposition 4}:}

The proof is straightforward by Lemma \ref{lemma1}.

\noindent \textbf{Proof of Theorem \ref{semantics inclusions}:}

1. Follows from Theorem \ref{Theorem 3}.

2. Follows from Theorem \ref{Theorem 3}.

3. See Definition \ref{complete}, Definition \ref{grounded} and Definition \ref{preferred}.

4. Let $F\! Lab_1$ be a semi-stable fuzzy labeling  of FAS $\mathcal{F}=\langle\mathcal{A},\mathcal{R}\rangle$.
If $F\! Lab_1$ is not preferred, then there exists a preferred fuzzy labeling $F\! Lab_2$ such that $F\! Lab_{2}^a$ strictly contains ${F\! Lab}_1^a$.
From Lemma \ref{lemma1}, it holds that ${F\! Lab}_2^r$ strictly contains ${F\! Lab}_1^r$.
It follows that ${F\! Lab}_1^u$ strictly contains ${F\!Lab}_2^u$ which contradicts that $F\! Lab_1$ is semi-stable.
Hence, every semi-stable fuzzy labeling is preferred.

5. See Definition \ref{semi-stable} and Definition \ref{stable}.

\noindent \textbf{Proof of Theorem \ref{correspondence relation for cac}:}

1. Suppose $E$ is a conflict-free fuzzy extension.
We now need to prove that $F\!Lab=Ext2F\!Lab(E)=\{E, E^{+}, (E\oplus E^{+})^c\}$ is a conflict-free fuzzy labeling. 
Since $E$ is conflict-free, it follows that all attacks in $E$ are tolerable, i.e., for any $(A,a),(B,b)\in E$, $b*\mathcal{R}_{_{BA}}+a\leq 1$.
Thus for any $A\in Args$, $\max_{B\in Att(A)} E(B)*\mathcal{R}_{_{BA}}+E(A)\leq 1$.
It directly follows that $F\!Lab$ satisfies TP.
Other conditions are easy to check.

2. Suppose $F\! Lab$ is a conflict-free fuzzy labeling.
We now need to prove that $E=F\! Lab2Ext(F\! Lab)={F\! Lab}^a$ is a conflict-free fuzzy extension.
Since $F\! Lab$ satisfies TP, it follows that for any $A\in Args$,  $\max_{B\in Att(A)} F\!Lab^{a}(B)*\mathcal{R}_{_{BA}}+F\!Lab^{a}(A)\leq 1$, i.e., $E(B)*\mathcal{R}_{_{BA}}+E(A)\leq 1$.
Thus, all attacks in $E$ are tolerable.
Therefore, $E$ is a conflict-free fuzzy extension.

3. Suppose $E$ is an admissible fuzzy extension.
We now need to prove that $F\!Lab=Ext2F\!Lab(E)=\{E, E^{+}, (E\oplus E^{+})^c\}$ is an admissible fuzzy labeling. 
Since $E$ is admissible, it follows that $E$ weakening defends every element in $E$.
Namely for each argument $A\in Args$, if $(B, \mathcal{A}(B))$ sufficiently attacks $(A, E(A))$, then there is an argument $C\in Att(B)$ and $(C, c)\in E$ s.t. $(C, c)$ weakens $(B, \mathcal{A}(B))$ to $(B, 1-E(A))$.
Thus there exists a fuzzy argument $(C,c)\in E$ s.t. $c*\mathcal{R}_{_{CB}}=E(A)$.
For this reason, we have $$E(A) =c*\mathcal{R}_{_{CB}}\leq\max_{X\in Att(B)}E(X)*\mathcal{R}_{_{XB}}=E^{+}(B)$$ by Definition \ref{the translation Ext2FLab}.
Therefore, for any argument $B\in Att(A)$, if $(B, \mathcal{A}(B))$ sufficiently attacks $(A, E(A))$, then  $E(A)\leq E^+(B)$, i.e., $F\!Lab^a(A)\leq F\!Lab^r(B)$ or $A^a\leq B^r$.
Consequently, $F\!Lab$ satisfies DP by Theorem \ref{Theorem 1}.
Other conditions are easy to check by Definition \ref{the translation Ext2FLab}.

4. Suppose $F\! Lab$ is an admissible fuzzy labeling.
We now need to prove that $E=F\! Lab2Ext(F\! Lab)={F\! Lab}^a$ is an admissible fuzzy extension.
It directly follows from point 2 that $E$ is conflict-free.
Then we prove that for an arbitrary element $(A, a)\in E$, it is weakening defended by $E$.
For each argument $B\in Att(A)$, if $(B,\mathcal{A}(B))$ sufficiently attacks $(A,a)$, then $(B,\mathcal{A}(B))$ sufficiently attacks $(A,E(A))$.
It directly follows from Theorem \ref{Theorem 1} that $E(A)=F\!Lab^a(A)\leq F\!Lab^r(B)$.
Thus, there exists at least an argument $C\in Att(B)$ and $(C,c)\in F\!Lab^a$ s.t. $c*\mathcal{R}_{_{CB}}=E(A)$ by WP.
Therefore, for any $(A, a)\in E$, if $(B,\mathcal{A}(B))$ sufficiently attacks $(A,a)$, then $\exists (C,c)\in E$ weakens $(B,\mathcal{A}(B))$ to $(B,1-E(A))$ which tolerably attacks $(A, a)$.
Hence, $E$ is an admissible fuzzy extension.

5. Suppose $E$ is a complete fuzzy extension.
We now need to prove that $F\!Lab=Ext2F\!Lab(E)=\{E, E^{+}, (E\oplus E^{+})^c\}$ is a complete fuzzy labeling. 
We first prove that $F\!Lab$ satisfies SDP.
Assume that for any argument $A\in Args$, $$A_d=\min\{\min_{X\in Att(A)}\{\max \{E^+(X),1-\mathcal{A}(X)*\mathcal{R}_{_{XA}}\}\},\mathcal{A}(A)\}.$$
By the proof of point 3, it follows that $F\!Lab$ is an admissible fuzzy labeling and thus for any $A\in Args$, $E(A)\leq A_d$ by DP and BP.
Then we need to prove that $E(A)\geq A_d$, i.e., $(A,A_d)\in E$.
According to the definition of complete fuzzy extension, we only need to prove that $(A,A_d)$ is weakening defended by $E$.
For any argument $B\in Att(A)$, if $(B,\mathcal{A}(B)) \text{ sufficiently attacks } (A,A_d)$, then $A_d> 1-\mathcal{A}(X)*\mathcal{R}_{_{XA}}$.
Since $A_d\leq \max \{E^+(B),1-\mathcal{A}(B)*\mathcal{R}_{_{BA}}\}$, we have $A_d\leq E^+(B)$.
Therefore, there exists an argument $C\in Att(B)$ and $(C,c)\in E$ s.t. $A_d= c*\mathcal{R}_{_{CB}}$.
Namely, $(C,c)$ weakens $(B,\mathcal{A}(B))$ to $(B,1-A_d)$ which tolerably attacks $(A,A_d)$.
It follows that $(A,A_d)$ is weakening defended by $E$.
Hence, $F\!Lab^a(A)=E(A)=A_d$.
Consequently, $F\!Lab$ satisfies SDP.
Other conditions are easy to check by Definition \ref{the translation Ext2FLab}.

6. Suppose $F\! Lab$ is a complete fuzzy labeling.
We now need to prove that $E=F\! Lab2Ext(F\! Lab)={F\! Lab}^a$ is a complete fuzzy extension.
It can be concluded that $E$ is an admissible fuzzy extension by the proof of point 4.
So we only need to prove that $E$ contains all the fuzzy arguments in $\mathcal{A}$ that it weakening defends.
Suppose $(A, a)\in \mathcal{A}$ is weakening defended by $E$.
It follows that if $(B, \mathcal{A}(B))$ sufficiently attacks $(A, a)$, then $E$ weakens $(B, \mathcal{A}(B))$ to be uncontroversial with $(A, a)$.
Namely there exists an argument $C\in Att(B)$ and $(C, c)\in E$ weakens $(B, \mathcal{A}(B))$ to $(B, 1-a)$, i.e., $c*\mathcal{R}_{_{CB}}=a$.
Since $(C, c)\in E={F\! Lab}^a$, we have $$a=c*\mathcal{R}_{_{CB}}\leq \max_{X\in Att(B)}{F\! Lab}^a*\mathcal{R}_{_{XB}}={F\! Lab}^r(B).$$
Thus for any $B\in Att(A)$ if $(B,\mathcal{A}(B))$ sufficiently attacks $(A, a)$, then $a\leq {F\! Lab}^r(B)$.
Let $S=\{B\in Args\mid (B,\mathcal{A}(B)) \text{ sufficiently attacks } (A,a)\}$.
It directly follows that 
\begin{align*}
   a&\leq \min\{\min_{B\in S}F\!Lab^r(B),\mathcal{A}(A)\}\\
   &\leq\min\{\min_{B\in S}\{\max \{F\!Lab^r(B),1-\mathcal{A}(B)*\mathcal{R}_{_{BA}}\}\},\mathcal{A}(A)\}. 
\end{align*}
And it is evident that 
\begin{align*}
   a&\leq \min\{\min_{B\notin S}\{1-\mathcal{A}(B)*\mathcal{R}_{_{BA}}\},\mathcal{A}(A)\}\\
   &\leq\min\{\min_{B\notin S}\{\max \{F\!Lab^r(B),1-\mathcal{A}(B)*\mathcal{R}_{_{BA}}\}\},\mathcal{A}(A)\}.
\end{align*}
It directly follows from SDP that 
$$a\leq\min\{\min_{B\in Att(A)}\{\max \{F\!Lab^r(B),1-\mathcal{A}(B)*\mathcal{R}_{_{BA}}\}\},\mathcal{A}(A)\}=E(A).$$
Therefore, we obtain $a\leq E(A)$, i.e., $(A,a)\in E$.
Consequently, $E$ contains all the fuzzy arguments in $\mathcal{A}$ that it weakening defends, i.e., $E$ is a complete fuzzy extension.

7. Suppose $E$ is a grounded fuzzy extension, i.e., it is a minimal complete fuzzy extension. From point 1 of Theorem \ref{Theorem 11}, complete fuzzy labelings and complete fuzzy extensions stand in a one-to-one relationship to each other. 
Therefore, $F\!Lab=Ext2F\!Lab(E)$ is a complete fuzzy labeling with minimal $F\!Lab^a$, i.e., it is a grounded fuzzy labeling.

8. Suppose $F\! Lab$ is a grounded fuzzy labeling, i.e., it is a complete fuzzy labeling with minimal $F\!Lab^a$.
From point 1 of Theorem \ref{Theorem 11}, complete fuzzy labelings and complete fuzzy extensions stand in a one-to-one relationship to each other.
Therefore, $E=F\! Lab2Ext(F\! Lab)$ is a minimal complete fuzzy extension, i.e., it is a grounded fuzzy extension.

9. Analogous to 7.

10. Analogous to 8.

\noindent \textbf{Proof of Theorem \ref{Theorem 11}:}

1. Under complete fuzzy labeling and complete fuzzy extension, we just need to prove that: when the domain and range of $F\!Lab2Ext$ are restricted to complete fuzzy labelings and complete fuzzy extensions, and the domain and range of $Ext2F\!Lab$ are restricted to complete fuzzy extensions and complete fuzzy labelings, $Ext2F\!Lab$ and $F\!Lab2Ext$ are bijective and each other's inverse.
So we only need to prove that $F\!Lab2Ext$ and $Ext2F\!Lab$ are each other's inverses since that implies that they are bijective.

From Definition \ref{the translation FLab2Ext} and Definition \ref{the translation Ext2FLab}, for each complete fuzzy extension $E$, $Ext2F\!Lab\big(Ext2F\!Lab(E)\big)=E$ is obvious.

Suppose $F\! Lab$ is a complete fuzzy labeling.
From Definition \ref{the translation FLab2Ext} and \ref{the translation Ext2FLab}, it follows that $F\!Lab2Ext(F\! Lab)={F\! Lab}^a$ and $Ext2F\!Lab\big({F\! Lab}^a\big)=\big({F\! Lab}^a$, $({{F\! Lab}^a})^{+}$, $\big({F\! Lab}^a\oplus ({{F\! Lab}^a})^{+}\big)^c$\big).
And therefore we only need to prove that $({{F\! Lab}^a})^{+}={F\! Lab}^r$.
\begin{itemize}
 \item For any $(A, a)\in {F\! Lab}^r$, there exists at least one attacker $(B, b)$ in ${F\! Lab}^a$ s.t. $b*\mathcal{R}_{_{BA}}=a$. Therefore it follows from Definition \ref{the translation Ext2FLab} that $(A, a)\in ({{F\! Lab}^a})^{+}$.
 \item For any $(A, a)\in ({{F\! Lab}^a})^{+}$, there exists at least one attacker $(B, b)$ in ${F\! Lab}^a$ s.t. $b*\mathcal{R}_{_{BA}}=a$. Therefore it follows from SWP that $(A, a)\in {F\! Lab}^r$.
\end{itemize}
Hence, $Ext2F\!Lab\big(F\! Lab2Ext(F\! Lab)\big)=F\!Lab$. 

2. Directly follows from point 1 and Theorem \ref{correspondence relation for cac}.

3. Directly follows from point 1 and Theorem \ref{correspondence relation for cac}.

\noindent \textbf{Proof of Theorem \ref{Theorem 12}:}

Let $\mathcal{F}=\langle\mathcal{A}, \mathcal{R}\rangle$ be the corresponding fuzzy argumentation system of $AF$ and let $F\!Lab$ be the corresponding fuzzy labeling of $Lab$.

1. Under conflict-free semantics, we need to prove that $F\!Lab$ is a conflict-free fuzzy labeling of $\mathcal{F}$.
Namely, for any argument $A\in Args$, 
\begin{align*}
A^{a}&\leq\min\{1-\max_{B\in Att(A)} B^{a}*\mathcal{R}_{_{BA}},\mathcal{A}(A)\}   &(\mbox{By BP,TP})\\
&=\min\{1-\max_{B\in Att(A)} B^{a},1\}   \\
&=1-\max_{B\in Att(A)} B^{a} \\
A^{u}&=1-A^{a}-A^{r}  &(\mbox{By UP})
\end{align*}

There are three cases:
\begin{itemize}
\setlength{\itemsep}{0pt}
\setlength{\parsep}{0pt}
\setlength{\parskip}{0pt}
    \item $Lab(A)=in$. 
According to Definition \ref{argumentlabeling}, we derive that there exists no argument $B\in Att(A)$ s.t. $Lab(B)=in$.
Thus for any argument $B\in Att(A)$, $B^a=0$.
From Definition \ref{correspond}, it follows that
\begin{align*}
A^{a}&=1\leq 1-\max_{B\in Att(A)} B^{a};\\
A^{r}&=0;\\
A^{u}&=0=1-A^{a}-A^{r}.
\end{align*}
\item $Lab(A)=out$. Trivial.
\item $Lab(A)=undec$. Trivial.
\end{itemize}
Therefore, $F\!Lab$ is a conflict-free fuzzy labeling.

2. Under admissible semantics, we need to prove that $F\!Lab$ is an admissible fuzzy labeling of $\mathcal{F}$.
Namely, for any argument $A\in Args$, 
\begin{align*}
A^{a}&\leq\min\{\min_{B\in Att(A)}\{\max \{B^{r},1-\mathcal{A}(B)*\mathcal{R}_{_{BA}}\}\},\mathcal{A}(A)\}   &(\mbox{By BP,DP})\\
&=\min\{\min_{B\in Att(A)}\{\max \{B^{r},0\}\},1\}\\
&=\min_{B\in Att(A)}B^{r} \\
A^{r}&\leq\max_{B\in Att(A)} B^{a}*\mathcal{R}_{_{BA}}=\max_{B\in Att(A)} B^{a}  &(\mbox{By WP)}\\
A^{u}&=1-A^{a}-A^{r}  &(\mbox{By UP})
\end{align*}

There are three cases:
\begin{itemize}
\setlength{\itemsep}{0pt}
\setlength{\parsep}{0pt}
\setlength{\parskip}{0pt}
    \item $Lab(A)=in$. 
According to Definition \ref{argumentlabeling}, we derive that $Lab(B)=out$ for any $B\in Att(A)$.
Thus for any argument $B\in Att(A)$, $B^a=0$ and $B^r=1$.
From Definition \ref{correspond}, it follows that
\begin{align*}
A^{a}&=1\leq\min_{B\in Att(A)}B^{r};\\
A^{r}&=0\leq\max_{B\in Att(A)} B^{a};\\
A^{u}&=0=1-A^{a}-A^{r}.
\end{align*}
\item $Lab(A)=out$.  
According to Definition \ref{argumentlabeling}, 
there exists at least one argument $B\in Att(A)$ and $Lab(B)=in$.
Thus $\exists B\in Att(A)$, $B^a=1$ and $B^r=0$.
From Definition \ref{correspond}, it follows that 
\begin{align*}
A^{a}&=0\leq\min_{B\in Att(A)} B^{r};\\
A^{r}&=1\leq\max_{B\in Att(A)} B^{a};\\
A^{u}&=0=1-A^{a}-A^{r}.
\end{align*}
\item $Lab(A)=undec$. Trivial.
\end{itemize}
Therefore, $F\!Lab$ is an admissible fuzzy labeling.

3. Under complete semantics, we need to prove that $F\!Lab$ is a complete fuzzy labeling of $\mathcal{F}$.
Namely, for any argument $A\in Args$, 
\begin{align*}
A^{a}&=\min\{\min_{B\in Att(A)}\{\max \{B^{r},1-\mathcal{A}(B)*\mathcal{R}_{_{BA}}\}\},\mathcal{A}(A)\}&(\mbox{By SDP})\\
&=\min\{\min_{B\in Att(A)}\{\max \{B^{r},0\}\},1\}\\
&=\min_{B\in Att(A)}B^{r}\\
A^{r}&=\max_{B\in Att(A)} B^{a}*\mathcal{R}_{_{BA}}=\max_{B\in Att(A)} B^{a} &(\mbox{By SWP})\\
A^{u}&=1-A^{a}-A^{r} &(\mbox{By UP})
\end{align*}

There are three cases:
\begin{itemize}
\setlength{\itemsep}{0pt}
\setlength{\parsep}{0pt}
\setlength{\parskip}{0pt}
    \item $Lab(A)=in$. 
According to Definition \ref{argumentlabeling}, we derive that $Lab(B)=out$ for any $B\in Att(A)$.
Thus for any argument $B\in Att(A)$, $B^a=0$ and $B^r=1$.
From Definition \ref{correspond}, it follows that
\begin{align*}
A^{a}&=1=\min_{B\in Att(A)} B^{r};\\
A^{r}&=0=\max_{B\in Att(A)} B^{a};\\
A^{u}&=0=1-A^{a}-A^{r}.
\end{align*}
\item $Lab(A)=out$.  
According to Definition \ref{argumentlabeling}, 
there exists at least one argument $B\in Att(A)$ and $Lab(B)=in$.
Thus $\exists B\in Att(A)$, $B^a=1$ and $B^r=0$.
From Definition \ref{correspond}, it follows that 
\begin{align*}
A^{a}&=0=\min_{B\in Att(A)} B^{r};\\
A^{r}&=1=\max_{B\in Att(A)} B^{a};\\
A^{u}&=0=1-A^{a}-A^{r}.
\end{align*}
    \item $Lab(A)=undec$. According to Definition \ref{argumentlabeling}, 
there exists no argument $B\in Att(A)$ s.t. $Lab(B)=in$, and there exists at least one argument $C\in Att(A)$ s.t. $Lab(C)=undec$.
Thus for any $B\in Att(A)$, $B^a=0$ and $\exists C\in Att(A)$ s.t. $C^r=0$.
From Definition \ref{correspond}, it follows that 
\begin{align*}
A^{a}&=0=\min_{B\in Att(A)} B^{r};\\
A^{r}&=0=\max_{B\in Att(A)} B^{a};\\
A^{u}&=1=1-A^{a}-A^{r}.
\end{align*}
\end{itemize}
Therefore, $F\!Lab$ is a complete fuzzy labeling.

4. Under grounded semantics, we prove it by the definitions of grounded extension and grounded fuzzy extension.
For the sake of simplicity, $S$ can represent either classical set or fuzzy set in the following proof.
In \cite{dung1995acceptability}, the grounded extension of AF is defined as the least fixed point of $F_{AF}$ where ${F}_{AF}$ is a function from the set of all the subsets of $Args$ to itself s.t. for any $S\subseteq Args$, 
$$F_{AF}(S)=\{A\in Args\vert (B,A)\in Att\Rightarrow\exists C\in S \text{ s.t. } (C,B)\in Att\}.$$
It means that for any argument $A\in F_{AF}(S)$, if $B$ attacks $A$, then $\exists C\in S$ attacks $B$.
In \cite{wu2016godel}, the grounded fuzzy extension of FAS $\mathcal{F}$ is defined as the least fixed point of $F_\mathcal{F}$, where $F_\mathcal{F}$ is a function from the set of all the subsets of $\mathcal{A}$ to itself, such that for any fuzzy set $S\subseteq \mathcal{A}$, 
$$F_\mathcal{F}(S)=\{(A,a)\vert S \text{ weakening defends } (A,a)\}.$$
It means that for any $(A,a)\in S$, if $(B,\mathcal{A}(B))$ sufficiently attacks $(A,a)$, then $\exists (C,c)\in S$ s.t. $(C,c)$ weakens $(B,\mathcal{A}(B))$ to $(B, 1-a)$.
It is easy to see that both $F_{AF}$ and $F_\mathcal{F}$ are monotonically increasing.
The grounded extension of AF is equal to $\bigcup_{i=1,...,\infty}F^i_{AF}(\varnothing)$ where $F^1_{AF}(\varnothing)=F_{AF}(\varnothing)$ and for $i>1$ $F^i_{AF}=F_{AF}(F^{i-1}_{AF}(\varnothing))$.
And the grounded fuzzy extension of $\mathcal{F}$ is equal to $\bigcup_{i=1,...,\infty}F^i_{\mathcal{F}}(\varnothing)$ where $F^1_{\mathcal{F}}(\varnothing)=F_{\mathcal{F}}(\varnothing)$ and for $i>1$ $F^i_{\mathcal{F}}=F_{\mathcal{F}}(F^{i-1}_{\mathcal{F}}(\varnothing))$.
We now prove that for any $A\in Args$ and $S\subseteq Args$, if $A\in F_{AF}(S)$, then $(A,1)\in F_\mathcal{F}(S)$.
Suppose $A\in F_{AF}(S)$. Then for any $(B,A)\in Att$, there is an argument $C\in S$ s.t. $(C,B)\in Att$.
Thus for any $(B,\mathcal{A}(B))$ (or $(B,1)$) sufficiently attacks $(A,1)$, $\exists (C,1)\in S$ s.t. $1*\mathcal{R}_{_{CB}}=1$, i.e., $(C,1)$ weakens $(B,\mathcal{A}(B))$ to $(B,0)$ which is tolerable attacks $(A,1)$, and therefore $(A,1)\in F_\mathcal{F}(S)$.
Accordingly, for any $i\in\{1,2,...\}$, if $A\in F^i_{AF}(\varnothing)$, then $(A,1)\in F^i_{\mathcal{F}}(\varnothing)$.
Consequently, if $A$ is an element of the grounded extension of $AF$, then $(A,1)$ is also an element of the grounded fuzzy extension of $\mathcal{F}$.

In \cite{baroni2011introduction,caminada2009logical}, it has been proved that the grounded extension and the grounded labeling stand in a one-to-one correspondence relationship.
If $A$ is an element of the grounded extension, then it is labeled as $in$ in the grounded labeling.
Similarly, the grounded fuzzy extension and the grounded fuzzy labeling stand in a one-to-one correspondence relationship by Theorem \ref{Theorem 11}.
Therefore, if $A$ is labeled as $in$ in the grounded labeling, then $A^a=1$ in the grounded fuzzy labeling.
Hence, the corresponding fuzzy labeling $F\!Lab$ is a complete fuzzy labeling with the least $F\!Lab^a$, i.e., $F\!Lab$ is the grounded fuzzy labeling of $\mathcal{F}$.

5. Under stable semantics, we need to prove that $F\!Lab$ is a stable fuzzy labeling of $\mathcal{F}$.
Suppose $Lab$ is stable, then $Lab$ is a complete labeling with $undec(Lab)=\varnothing$.
By the proof of point 3, $F\!Lab$ is a complete fuzzy labeling of $\mathcal{F}$ and $F\!Lab^u=\varnothing$.
Therefore, $F\!Lab$ is a stable fuzzy labeling.

%
%
%
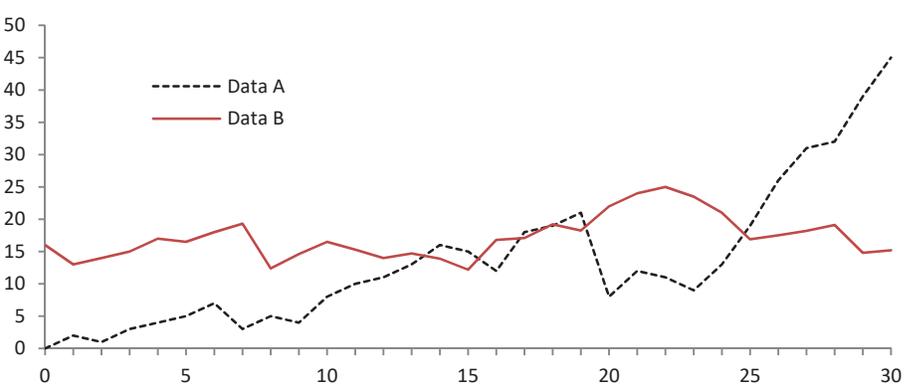
\end{document}